\documentclass[conference]{IEEEtran}

\makeatletter
\def\ps@headings{%
\def\@oddhead{\mbox{}\scriptsize\rightmark \hfil \thepage}%
\def\@evenhead{\scriptsize\thepage \hfil \leftmark\mbox{}}%
\def\@oddfoot{}%
\def\@evenfoot{}}
\makeatother 

\usepackage{cite}      
\usepackage{subfigure} 
\usepackage{url}       
\usepackage{algorithm}
\usepackage{array}
\usepackage{amsmath,amsfonts,amssymb,graphicx}
\usepackage{bm}
\usepackage{mathrsfs}
\usepackage{multicol}
\usepackage[end]{algpseudocode}
\usepackage{multirow}
\usepackage{dsfont}

\usepackage{subfigure}

\newtheorem{theorem}{\textbf{Theorem}}

\newtheorem{corollary}{\textit{Corollary}}

\newtheorem{definition}{Definition}

\newcommand {\xA} {\mathcal{A}}

\newcommand{\commnt}[1] {$//$ \textsc{#1} }
\newcommand{\xO} {\mathcal{O}}
\newcommand{\hX} {\hat{\theta}}
\newcommand{\bX} {\hat{\overline{\theta}}}
\newcommand{\xom} {o^*_m}
\newcommand{\xoh} {o^*_h}
\newcommand{\aom} {o^*_m}

\newcommand{\kom} {K^*_m}

\begin{document}
\title{Decentralized Online Learning Algorithms for Opportunistic Spectrum Access}

\author{\IEEEauthorblockN{Yi Gai and Bhaskar Krishnamachari}
\IEEEauthorblockA{Ming Hsieh Department of Electrical Engineering\\
University of Southern California\\
Los Angeles, CA 90089, USA\\
Email: $\{$ygai, bkrishna$\}$@usc.edu}}

\maketitle

\begin{abstract}

The fundamental problem of multiple secondary users contending for
opportunistic spectrum access over multiple channels in cognitive
radio networks has been formulated recently as a decentralized
multi-armed bandit (D-MAB) problem. In a D-MAB problem there are $M$ users
and $N$ arms (channels) that each offer i.i.d. stochastic rewards with unknown means
so long as they are accessed without collision.
The goal is to design a decentralized online learning policy that incurs minimal regret,
defined as the difference between the total expected rewards accumulated by a model-aware genie, and that
obtained by all users applying the policy. We make two contributions in this paper.
First, we consider the setting where the users have a prioritized ranking, such that it is
desired for the $K$-th-ranked user to learn to access the arm offering the $K$-th highest mean reward.
For this problem, we present the first distributed policy that yields regret that is
uniformly logarithmic over time without requiring any prior assumption about the mean rewards. Second,
we consider the case when a fair access policy is required, i.e., it is desired for all users to experience
the same mean reward. For this problem, we present a distributed policy that yields order-optimal regret scaling
with respect to the number of users and arms, better than previously proposed policies in the literature. Both of our
distributed policies make use of an innovative modification of the well known UCB1 policy for the
classic multi-armed bandit problem that allows a single user to learn how to play the arm that yields
the $K$-th largest mean reward.
\end{abstract}

\section{Introduction}\label{sec:intro}

Developing dynamic spectrum access mechanisms to enable more
efficient spectrum utilization is one of the most challenging issues
in cognitive radio systems \cite{Yucek:2009}. In this paper, we
focus on a problem of opportunistic spectrum access in cognitive
radio networks, where at every time slot, each of the $M$
decentralized secondary users searches for idle channels which are
not occupied by primary users temporarily among $N \geq M$ channels. We
assume that the throughput of these $N$ channels evolves i.i.d. over
time with any arbitrary, bounded-support distribution, which is
unknown to the users. These distributed players can only learn from
their local observations and collide (with reward penalty) when
choosing the same arm. The desired objective is to develop a
sequential policy running at each user to make a selection among
multiple choices, where there is no information exchange, such that
the sum-throughput of all distributed users is maximized, assuming
an interference model whereby at most one secondary user can derive
benefit from any channel.

Multi-Armed Bandit problem (MAB,
see~\cite{Lai:Robbins,Anantharam,Anantharam:1987,Auer:2002,Gai:2010})
is a fundamental mathematical framework for learning the unknown
variables. In its simplest form of classic non-Bayesian version
studied by Lai and Robbins~\cite{Lai:Robbins}, there are $N$ arms,
each providing stochastic rewards that are independent and
identically distributed over time, with unknown means. A policy is
desired to pick one arm at each time sequentially, to maximize the
reward. Anantharam \emph{et al.}~\cite{Anantharam} extend this work
to the case when $M$ simultaneous plays are allowed, with
centralized scheduling of the players.

A fundamental tradeoff between exploration and exploitation is
captured by MAB problems: on the one hand, various arms should be
explored often enough in order to learn their parameters, and on the
other hand, the prior observations should be exploited to gain the
best possible immediate rewards. A key metric in evaluating a given
policy for this problem is \emph{regret}, which is defined as the
difference between the expected reward gained by a \emph{prior} that
always makes the optimal choice and that obtain by the given policy.
The regret achieved by a policy can be evaluated in terms of its
growth over time. Many of the prior works on multi-armed bandits
show logarithmic scaling of the regret over time.

While most of the prior work on MAB focused on the centralized
policies, motivated by the problem of opportunistic access in cognitive
radio networks, Liu and Zhao~\cite{Liu:Zhao, Liu:zhao:2010}, and
Anandkumar \emph{et al.}~\cite{Anandkumar:Infocom:2010, Anandkumar:JSAC} have both
developed policies for the problem of $M$ distributed players
operating $N$ independent arms. There are two problem formulations of interest when considering
distributed MAB: a) the \emph{prioritized access problem}, where it is desired to prioritize a ranked set of users so
that the $K$-th ranked user learns to access the arm with the $K$-th highest
reward, and b) the \emph{fair access problem}, where the goal is to ensure that each user receives
the same reward in expectation. For the prioritized access problem, Anandkumar \emph{et al.}~\cite{Anandkumar:Infocom:2010}
present a distributed policy that yields regret that is logarithmic in time, but requires
prior knowledge of the arm reward means. For the fair access problem, they propose in~\cite{Anandkumar:Infocom:2010, Anandkumar:JSAC}
a randomized distributed policy that is logarithmic with respect to time and scales as
$O(M^2 N)$ with respect to the number of arms and users. Liu and Zhao~\cite{Liu:Zhao, Liu:zhao:2010} also
treat the fair access problem and present the TDFS policy which yields
asymptotically logarithmic regret with respect to time and scales as $O(M(\max\{M^2, (N-M)M\}))$
with respect to the number of arms and users.

In this paper we make significant new contributions to both problem
formulations. For the prioritized access problem, we present a
distributed learning policy DLP that results in a regret that is
uniformly logarithmic in time and, unlike the prior work
in~\cite{Anandkumar:Infocom:2010}, does not require any prior
knowledge about the arm reward means. For the fair access problem,
we present another distributed learning policy DLF, which yields
regret that is also uniformly logarithmic in time and that scales as
$O(M(N-M))$ with respect to the number of users $M$ and the number
of arms $N$. As it has been shown in~\cite{Liu:zhao:2010} that the
lower-bound of regret for distributed policies also scales as
$\Omega(M(N-M))$, this is not only a better scaling than the
previous state of the art, it is, in fact, order-optimal.

A key subroutine of both decentralized learning policies running at
each user involves selecting an arm with the desired rank order of
the mean reward. For this, we present a new policy that we refer to
as SL($K$), which is a non-trivial generalization of UCB1 in
\cite{Auer:2002}. SL($K$) provides a general solution for selecting
an arm with the $K$-th largest expected rewards for classic MAB
problems with $N$ arms.

This paper is organized as follows. We present in
section~\ref{sec:formulation} the problem formulation. In section
\ref{sec:kth}, we first present our SL($K$) policy, which is a
general policy to play an arm with $K$-th largest expected reward
for classic multi-armed bandits, and then present our decentralized
DLP policy in section \ref{sec:DLP} and DLF policy in section
\ref{sec:DLF} based on SL($K$) policy. Both policies are
polynomial-storage polynomial-time-per-step learning policies. We
show that the regrets of all policies we proposed are logarithmic in
time and polynomial in the number of users and channels, and we
compare the upper bound of the regrets of different policies. In
section \ref{sec:simulation}, we compare the decentralized learning
policies with simulation results. Finally, section
\ref{sec:conclusion} concludes the paper.

\section{Problem Formulation}\label{sec:formulation}

We consider a cognitive system with $N$ channels (arms) and $M$
decentralized secondary users (players). The throughput of $N$
channels are defined by random processes $X_i(n)$, $1 \leq i \leq
N$. Time is slotted and denoted by the index $n$. We assume that
$X_i(n)$ evolves as an i.i.d. random process over time, with the
only restriction that its distribution have a finite support.
Without loss of generality, we normalize $X_i(n) \in [0,1]$. We do
not require that $X_i(n)$ be independent across $i$. This random
process is assumed to have a mean $\theta_{i} = E[X_i]$, that is
unknown to the users and distinct from each other. We denote the set
of all these means as $\Theta = \{\theta_{i}, 1\leq i \leq N\}$.

At each decision period $n$ (also referred to interchangeably as
time slot), each of the $M$ decentralized users selects an arm only
based on its own observation histories under a decentralized policy.
When a particular arm $i$ is selected by user $j$, the value of
$X_i(n)$ is only observed by user $j$, and if there is no other user
playing the same arm, a reward of $X_i(n)$ is obtained. Else, if
there are multiple users playing the same arm, then we assume that,
due to collision, at most one of the conflicting users $j'$ gets
reward $X_i(n)$, while the other users get zero reward. This
interference assumption covers practical models in networking
research, such as the perfect collision model (in which none of the
conflicting users derive any benefit) and CSMA with perfect sensing
(in which exactly one of the conflicting user derives benefit from
the channel). We denote the first model as $\mathbf{M}_1$ and the
second model as $\mathbf{M}_2$.

We denote the decentralized policy for user $j$ at time $n$ as
$\pi_j(n)$, and the set of policies for all users as $\mathcal{\pi}
= \{\pi_j(n), 1 \leq j \leq M \} $. We are interested in designing
decentralized policies, under which there is no information exchange
among users, and analyze them with respect to \emph{regret}, which
is defined as the gap between the expected reward that could be
obtained by a genie-aided perfect selection and that obtained by the
policy. We denote $\xO_M^*$ as a set of $M$ arms with $M$ largest
expected rewards. The regret can be expressed as:
\begin{equation}
\mathfrak{R}^\pi (\Theta;n)  = n\sum\limits_{i \in \xO^*_M} \theta_i
- E^\pi[\sum_{t = 1}^n S_{\pi(t)}(t)
 ]
\end{equation}
where $S_{\pi(t)}(t)$ is the sum of the actual reward obtained by
all users at time $t$ under policy $\pi(t)$, which could be
expressed as:
\begin{equation}
  S_{\pi(t)}(t) =  \sum\limits_{i =
1}^N\sum\limits_{j = 1}^M X_i(t) \mathds{I}_{i,j}(t),
\end{equation}
where for $\mathbf{M}_1$, $\mathds{I}_{i,j}(t)$ is defined to be 1
if user $j$ is the only user to play arm $i$, and 0 otherwise; for
$\mathbf{M}_2$, $\mathds{I}_{i,j}(t)$ is defined to be 1 if user $j$
is the one with the smallest index among all users playing arm $i$
at time $t$, and 0 otherwise. Then, if we denote $V_{i,j}^\pi(n) = E
[\sum_{t = 1}^n \mathds{I}_{i,j}(t)]$, we have:
\begin{equation}
E^\pi[\sum_{t = 1}^n S_{\pi(t)}(t)] = \sum\limits_{i =
1}^N\sum\limits_{j = 1}^M \theta_i E[V_{i,j}^\pi(n)]
\end{equation}

Besides getting low total regret, there could be other system
objectives for a given D-MAB. We consider two in this paper. In the
prioritized access problem, we assume that each user has information
of a distinct allocation order. Without loss of generality, we
assume that the users are ranked in such a way that the $m$-th user
seeks to access the arm with the $m$-th highest mean reward. In the
fair access problem, users are treated equally to receive the same
expected reward.


\section{Selective Learning of the $K$-th Largest Expected Reward}
\label{sec:kth}

We first propose a general policy to play an arm with the $K$-th
largest expected reward ($1\leq K \leq N$) for classic multi-armed
bandit problem with $N$ arms and one user, since the key idea of our
proposed decentralized policies running at each user in section
\ref{sec:DLP} and \ref{sec:DLF} is that user $m$ will run a learning
policy targeting an arm with $m$-th largest expected reward.

Our proposed policy of learning an arm with $K$-th largest expected
reward is shown in Algorithm \ref{alg:kthLargest}.

{\renewcommand\baselinestretch{1.0}
\begin{algorithm} [ht]
\caption{Selective learning of the $K$-th largest expected rewards
(SL($K$))} \label{alg:kthLargest}

\begin{algorithmic}[1]
\State \commnt{ Initialization}

\For {$t = 1$ to $N$}
    \State Let $i = t$ and play arm $i$;
    \State $\hat{\theta}_{i}(t) = X_i(t)$;
    \State $n_{i}(t) = 1$;
\EndFor

\State \commnt{Main loop}

\While {1}
    \State $t = t + 1$;
    \State Let the set $\xO_K$ contains the $K$ arms with the
    $K$ largest values in (\ref{equ:d01})
    \begin{equation}
    \label{equ:d01}
    \hX_i(t-1) + \sqrt{\frac{2 \ln t}{n_i(t-1)}};
    \end{equation}
    \State Play arm $k$ in $\xO_K$ such that
    \begin{equation}
    \label{equ:d02}
    k = \arg\min_{i \in \xO_K} \hX_i(t-1) - \sqrt{\frac{2 \ln t}{n_i(t-1)}};
    \end{equation}
    \State $\hX_{k}(t) = \frac{\hX_{k}(t-1) n_{k}(t-1) + X_{k}(t)}{ n_{k}(t-1) +1}$;
    \State $n_{k}(t) = n_{k}(t-1) + 1$;
\EndWhile
\end{algorithmic}
\end{algorithm}
\par}

We use two $1$ by $N$ vectors to store the information after we play
an arm at each time slot. One is $(\hat{ \theta }_{i})_{1 \times N}$
in which $\hat{ \theta }_{i}$ is the average (sample mean) of all
the observed values of $X_i$ up to the current time slot (obtained
through potentially different sets of arms over time). The other one
is $(n_{i})_{1 \times N}$ in which $n_{i}$ is the number of times
that $X_i$ has been observed up to the current time slot.

Note that while we indicate the time index in Algorithm
\ref{alg:kthLargest} for notational clarity, it is not necessary to
store the matrices from previous time steps while running the
algorithm. So SL($K$) policy requires storage linear in $N$.

\emph{\textbf{Remark}}: SL($K$) policy generalizes UCB1 in
\cite{Auer:2002} and presents a general way to pick an arm with the
$K$-th largest expected rewards for a classic multi-armed bandit
problem with $N$ arms (without the requirement of distinct expected
rewards for different arms).

Now we present the analysis of the upper bound of regret, and show
that it is linear in $N$ and logarithmic in time. We denote $\xA_K$
as the set of arms with $K$-th largest expected reward. Note that
Algorithm \ref{alg:kthLargest} is a general algorithm for picking an
arm with the K-th largest expected reward for the classic
multi-armed bandit problems, where we allow multiple arms with
$K$-th largest expected reward, and all these arms retreated as
optimal arms. The following theorem holds for Algorithm
\ref{alg:kthLargest}.

\theorem \label{theorem:kth} Under the policy specified in Algorithm
\ref{alg:kthLargest}, the expected number of times that we pick any
arm $i \notin \xA_K$ after $n$ time slots is at most:
\begin{equation}
 \frac{8 \ln n}{\Delta_{K,i}} + 1 + \frac{2 \pi^2}{3}.
\end{equation}
where $\Delta_{K,i} = |\theta_K - \theta_i|$, $\theta_K$ is the
$K$-th largest expected reward.

\begin{IEEEproof}
Denote $T_i(n)$ as the number of times that we pick arm $i \notin
\xA_K$ at time $n$. Denote $C_{t, n_i}$ as $\sqrt{ \frac{ (L+1) \ln
t }{ n_i} }$. Denote $\bX_{i, n_i}$ as the average (sample mean) of
all the observed values of $X_i$ when it is observed $n_i$ time.
$\xO_K^*$ is denoted as the set of $K$ arms with $K$ largest
expected rewards.

Denote by $I_i(n)$ the indicator function which is equal to $1$ if
$T_i(n)$ is added by one at time $n$. Let $l$ be an arbitrary
positive integer. Then, for any arm $i$ which is not a desired arm,
i.e., $i \notin \xA_K$:
\begin{equation} \label{equ:d002}
\begin{split}
 T_i(n) & = 1 + \sum\limits_{t = N+1}^n \mathds{1} \{ I_i(t)\} \\
 & \leq l + \sum\limits_{t = N+1}^n \mathds{1} \{ I_i(t) , T_i(t-1) \geq l \} \\
 & \leq l + \sum\limits_{t = N+1}^n \left( \mathds{1} \{ I_i(t) , \theta_i < \theta_K, T_i(t-1) \geq l \} \right. \\
 & \quad\quad\quad\quad + \left. \mathds{1} \{ I_i(t) , \theta_i > \theta_K, T_i(t-1) \geq l
 \} \right)
\end{split}
\end{equation}
where $\mathds{1}(x)$ is the indicator function defined to be 1 when
the predicate $x$ is true, and 0 when it is false.

Note that for the case $\theta_i < \theta_K$, arm $i$ is picked at
time $t$ means that there exists an arm $j(t) \in \xO^*_K$, such
that $j(t) \notin \xO_K$. This means the following inequality holds:
\begin{equation} \label{equ:d03}
 \bX_{j(t), T_{j(t)}(t-1)} + C_{t-1, T_{j(t)}(t-1)} \leq \bX_{i, T_i(t-1) + C_{t-1,
 T_i(t-1)}}.
\end{equation}

Then, we have
\begin{equation}
\begin{split}
& \sum\limits_{t = N+1}^n \mathds{1} \{ I_i(t) , \theta_i <
\theta_K, T_i(t-1) \geq l \} \\
& \leq \sum\limits_{t = N+1}^n\mathds{1} \{ \bX_{j(t), T_{j(t)}(t-1)} + C_{t-1, T_{j(t)}(t-1)}\\
    & \quad\quad\quad \leq \bX_{i, T_i(t-1)} + C_{t-1,
 T_i(t-1)}, T_i(t-1) \geq l \}\\
 & \leq \sum\limits_{t = N+1}^n\mathds{1} \{ \min\limits_{0 <
n_{j(t)} < t} \bX_{j(t), n_{j(t)}} + C_{t-1, n_{j(t)}} \\
   & \quad\quad\quad\quad\quad\quad \leq \max\limits_{l \leq n_i <t} \bX_{i, n_i} + C_{t-1,
 n_i} \}\\
 & \leq \sum\limits_{t = 1}^{\infty} \sum\limits_{n_{j(t)} = 1}^{t-1} \sum\limits_{n_i = l}^{t-1}  \mathds{1} \{
    \bX_{j(t), n_{j(t)}} + C_{t, n_{j(t)}} \leq \bX_{i, n_i}+ C_{t, n_i} \}
\end{split}
\end{equation}

$\bX_{j(t), n_{j(t)}} + C_{t, n_{j(t)}} \leq \bX_{i, n_i}+ C_{t,
n_i}$ implies that at least one of the following must be true:
\begin{equation}
\label{equ:de1}
 \bX_{j(t), n_{j(t)}} \leq \theta_{j(t)} - C_{t, n_{j(t)}},
\end{equation}
\begin{equation}
\label{equ:de2}
 \bX_{i, n_i} \geq \theta_{i} + C_{t,n_i},
\end{equation}
\begin{equation}
\label{equ:de3}
 \theta_{j(t)} < \theta_{i}  + 2 C_{t,n_i}.
\end{equation}
Applying the Chernoff-Hoeffding bound~\cite{Pollard}, we could find
the upper bound of (\ref{equ:de1}) and (\ref{equ:de2}) as,
\begin{equation}
  Pr\{\bX_{j(t), n_{j(t)}} \leq \theta_{j(t)} - C_{t, n_{j(t)}} \} \leq e^{-4 \ln t} =
 t^{-4},
\end{equation}
\begin{equation}\label{equ:f01}
  Pr\{\bX_{i, n_i} \geq \theta_{i} + C_{t,n_i} \} \leq e^{-4 \ln t} = t^{-4}
\end{equation}
For $l \geq \left\lceil \frac{8 \ln n}{\Delta_{K,i}^2}
\right\rceil$,
\begin{equation}
 \label{equ:p04}
\begin{split}
 & \theta_{j(t)} - \theta_{i}  - 2 C_{t,n_i}  \\
 & \geq \theta_{K} - \theta_{i} - 2 \sqrt{ \frac{ 2 \Delta_{K,i}^2 \ln t }{ 8  \ln n } } \\
 & \geq \theta_{K} - \theta_{i} - \Delta_{K,i} = 0,
\end{split}
\end{equation}
so (\ref{equ:de3}) is false when $l \geq \left\lceil \frac{8 \ln
n}{\Delta_{K,i}^2} \right\rceil$.

Note that for the case $\theta_i > \theta_K$, when arm $i$ is picked
at time $t$, there are two possibilities: either $\xO_K = \xO^*_K$,
or $\xO_K \neq \xO^*_K$. If $\xO_K = \xO^*_K$, the following
inequality holds:
\begin{equation}
 \bX_{i, T_i(t-1)} - C_{t-1, T_i(t-1)} \leq \bX_{K, T_{K}(t-1)} - C_{t-1,
 T_{K}(t-1)}. \nonumber
\end{equation}
If $\xO_K \neq \xO^*_K$, $\xO_K$ has at least one arm $h(t) \notin
\xO^*_{K}$. Then, we have:
\begin{equation}
 \bX_{i, T_i(t-1)} - C_{t-1, T_i(t-1)} \leq \bX_{h(t), T_{h(t)}(t-1)} - C_{t-1,
 T_{h(t)}(t-1)}. \nonumber
\end{equation}
So to conclude both possibilities for the case $\theta_i >
\theta_K$, if we denote $\xO^*_{K-1} = \xO^*_{K}-\xA_K$, at each
time $t$ when arm $i$ is picked, these exists an arm $h(t) \notin
\xO^*_{K-1}$, such that
\begin{equation} \label{equ:d003}
 \bX_{i, T_i(t-1)} - C_{t-1, T_i(t-1)} \leq \bX_{h(t), T_{h(t)}(t-1)} - C_{t-1,
 T_{h(t)}(t-1)}.
\end{equation}

Then similarly, we can have:
\begin{equation}
\begin{split}
& \sum\limits_{t = N+1}^n \mathds{1} \{ I_i(t) , \theta_i > \theta_K, T_i(t-1) \geq l \} \\
 & \leq \sum\limits_{t = 1}^{\infty} \sum\limits_{n_i = l}^{t-1} \sum\limits_{n_{h(t)} = 1}^{t-1}   \mathds{1} \{
    \bX_{i, n_i}- C_{t, n_i}  \leq \bX_{h(t), n_{h(t)}} - C_{t, n_{h(t)}} \}
\end{split}
\end{equation}

Note that $\bX_{i, n_i}- C_{t, n_i}  \leq \bX_{h(t), n_{h(t)}} -
C_{t, n_{h(t)}}$ implies one of the following must be true:
\begin{equation}
\label{equ:de4}
 \bX_{i, n_i} \leq \theta_{i} - C_{t, n_i},
\end{equation}
\begin{equation}
\label{equ:de5}
 \bX_{h(t), n_{h(t)}} \geq \theta_{h(t)} +C_{t, n_{h(t)}},
\end{equation}
\begin{equation}
\label{equ:de6}
 \theta_{i} < \theta_{h(t)} + 2 C_{t, n_i}.
\end{equation}

We again apply the Chernoff-Hoeffding bound and get $Pr\{\bX_{i,
n_i} \leq \theta_{i} - C_{t, n_i}\} \leq t^{-4}$, $Pr\{\bX_{h(t),
n_{h(t)}} \geq \theta_{h(t)} +C_{t, n_{h(t)}}\} \leq t^{-4}$.

Also note that for $l \geq \left\lceil \frac{8 \ln
n}{\Delta_{K,i}^2} \right\rceil$,
\begin{equation}
 \label{equ:p4}
\begin{split}
 & \theta_{i} - \theta_{h(t)}  - 2 C_{t, n_i}  \\
 & \geq \theta_{i} - \theta_{K} - \Delta_{K,i} \geq 0,
\end{split}
\end{equation}
so (\ref{equ:de6}) is false.

Hence, we have
\begin{equation}
\begin{split}
 & \mathbb{E}[T_i(n)]
  \leq \left\lceil \frac{8 \ln n}{\Delta_{K,i}^2}
   \right\rceil + \sum\limits_{t = 1}^{\infty} \sum\limits_{n_{j(t)} = 1}^{t-1} \sum\limits_{n_i = \left\lceil (8 \ln n)/
   \Delta_{K,i}^2
   \right\rceil}^{t-1} \\
   & (Pr\{\bX_{j(t), n_{j(t)}} \leq \theta_{j(t)} - C_{t, n_{j(t)}} \}  + Pr\{\bX_{i, n_i} \geq \theta_{i} + C_{t,n_i} \})\\
   & + \sum\limits_{t = 1}^{\infty} \sum\limits_{n_i = \left\lceil (8 \ln n)/
   \Delta_{K,i}^2
   \right\rceil}^{t-1} \sum\limits_{n_{h(t)} = 1}^{t-1} \\
   & ( Pr\{\bX_{i,
n_i} \leq \theta_{i} - C_{t, n_i}\} + Pr\{\bX_{h(t),
n_{h(t)}} \geq \theta_{h(t)} +C_{t, n_{h(t)}}\} )\\
& \leq \frac{8 \ln n}{\Delta_{K,i}^2} + 1 + 2 \sum\limits_{t =
1}^{\infty}
 \sum\limits_{n_{j(t)} = 1}^{t-1}  \sum\limits_{n_i = 1}^{t-1} 2
 t^{-4} \\
 & \leq \frac{8 \ln n}{\Delta_{K,i}^2} + 1 + \frac{2 \pi^2}{3}.
\end{split}
\end{equation}

\end{IEEEproof}

The definition of \emph{regret} for the above problem is different
from the traditional multi-armed bandit problem with the goal of
maximization or minimization, since our goal now is to pick the arm
with the $K$-th largest expected reward and we wish we could
minimize the number of times that we pick the wrong arm. Here we
give two definitions of the regret to evaluate the SL($K$) policy.

\begin{definition} We define the \emph{regret of type 1 at each time slot} as the absolute difference
between the expected reward that could be obtained by a genie that
can pick an arm with $K$-th largest expected reward, and that
obtained by the given policy at each time slot. Then the \emph{total
regret of type 1 by time $n$} is defined as sum of the regret at
each time slot, which is:
\begin{equation}
\begin{split}
 \mathfrak{R}^\pi_1 (\Theta;n) & = \sum_{t = 1}^n |\theta_K  - E^\pi[S_{\pi(t)}(t)
 ]|
 \end{split}
\end{equation}

\end{definition}

\begin{definition} We define the \emph{total regret of type 2 by time $n$} as the absolute
difference between the expected reward that could be obtained by a
genie that can pick an arm with $K$-th largest expected reward, and
that obtained by the given policy after $n$ plays, which is:
\begin{equation}
\mathfrak{R}^\pi_2 (\Theta;n)  = |n\theta_K  - E^\pi[\sum_{t = 1}^n
S_{\pi(t)}(t)
 ]|
\end{equation}
\end{definition}

Here we note that $\forall n$, $\mathfrak{R}^\pi_2 (\Theta;n) \leq
\mathfrak{R}^\pi_1 (\Theta;n)$ because $|n\theta_K  - E^\pi[\sum_{t
= 1}^n S_{\pi(t)}(t)
 ]| = |n\theta_K  - \sum_{t = 1}^n E^\pi[
S_{\pi(t)}(t)
 ]| \leq \sum_{t = 1}^n |\theta_K  - E^\pi[S_{\pi(t)}(t)
 ]|$.


\corollary \label{corollary:kth} The expected regret under both
definitions is at most
\begin{equation}
\sum\limits_{i: i \notin \xA_k} (\frac{8 \ln
 n}{\Delta_{K,i}}) + (1 + \frac{2 \pi^2}{3}) \sum\limits_{i: i \notin
 \xA_k} \Delta_{K,i}.
\end{equation}

\begin{IEEEproof} Under the SL($K$) policy, we have:
\begin{equation}
 \label{equ:d06}
\begin{split}
 & \mathfrak{R}^\pi_2 (\Theta;n)  \leq \mathfrak{R}^\pi_1
(\Theta;n) \\
 & = \sum_{t = 1}^n |\theta_K  - E^\pi[S_{\pi(t)}(t)
 ]|\\
& = \sum\limits_{i: i \notin \xA_k} \Delta_{K,i} \mathbb{E}[T_i(n)]\\
 & \leq  \sum\limits_{i: i \notin \xA_k} (\frac{8 \ln
 n}{\Delta_{K,i}}) + (1 + \frac{2 \pi^2}{3}) \sum\limits_{i: i \notin
 \xA_k} \Delta_{K,i}.
\end{split}
\end{equation}
\end{IEEEproof}

Corollary \ref{corollary:kth} shows the upper bound of the regret of
SL($K$) policy. It grows logarithmical in time and linearly in the
number of arms.

\section{Distributed Learning with Prioritization} \label{sec:DLP}

We now consider the distributed multi-armed bandit problem with
prioritized access. Our proposed decentralized policy for $N$ arms
with $M$ users is shown in Algorithm \ref{alg:distributed}.

\begin{algorithm} [ht]
\caption{Distributed Learning Algorithm with Prioritization for $N$
Arms with $M$ Users Running at User $m$ (DLP)}
\label{alg:distributed}

\begin{algorithmic}[1]
\State \commnt{ Initialization}

\For {$t = 1$ to $N$}
    \State Play arm $k$ such that $k = ((m+t) \mod N) + 1$;
    \State $\hat{\theta}_{k}^m(t) = X_k(t)$;
    \State $n_{k}^m(t) = 1$;
\EndFor

\State \commnt{Main loop}

\While {1}
    \State $t = t + 1$;
    \State Play an arm $k$ according to policy SL($m$) specified in Algorithm \ref{alg:kthLargest};
    \State $\hat{\theta}_{k}^m(t) = \frac{\hat{\theta}_{k}^m(t-1) n_{k}^m(t-1) + X_{k}(t)}{ n_{k}^m(t-1) +1}$;
    \State $n_{k}^m(t) = n_{k}^m(t-1) + 1$;
\EndWhile
\end{algorithmic}
\end{algorithm}

In the above algorithm, line \ref{a3:line:2} to \ref{a3:line:6} is
the initialization part, for which user $m$ will play each arm once
to have the initial value in $(\hat{\theta }_{i}^m)_{1 \times N}$
and $(n_{i}^m)_{1 \times N}$. Line \ref{a3:line:3} ensures that
there will be no collisions among users. Similar as in Algorithm
\ref{alg:kthLargest}, we indicate the time index for notational
clarity. Only two $1$ by $N$ vectors, $(\hat{ \theta }_{i}^m)_{1
\times N}$ and $(n_{i}^m)_{1 \times N}$, are used by user $m$ to
store the information after we play an arm at each time slot.

We denote $\xom$ as the index of arm with the $m$-th largest
expected reward. Note that $\{\xom\}_{1 \leq m \leq M} = \xO_M^*$.
Denote $\Delta_{i,j} = |\theta_i - \theta_j|$ for arm $i$, $j$. We
now state the main theorem of this section.

\theorem \label{theorem:distributed} The expected regret under the
DLP policy specified in Algorithm \ref{alg:distributed} is at most
\begin{equation}
\begin{split}
& \sum_{m = 1}^M \sum\limits_{i \neq \xom}
 (\frac{8 \ln n}{\Delta_{\xom, i}^2} + 1 + \frac{2
 \pi^2}{3}) \theta_{\aom} \\
 &+ \sum_{m = 1}^M \sum\limits_{h \neq m} (\frac{8 \ln n}{\Delta_{\xoh, \xom}^2} + 1 + \frac{2
 \pi^2}{3}) \theta_{\aom}.
\end{split}
\end{equation}

\begin{IEEEproof} Denote $T_{i,m}(n)$ the number
of times that user $m$ pick arm $i$ at time $n$.

For each user $m$, the regret under DLP policy can arise due to two
possibilities: (1) user $m$ plays an arm $i \neq \aom$; (2) other
user $h \neq m$ plays arm $\aom$. In both cases, collisions may
happen, resulting a loss which is at most $\theta_{\aom}$.
Considering these two possibilities, the regret of user $m$ is upper
bounded by:
\begin{equation}
 \mathfrak{R}^\pi (\Theta,m;n) \leq \sum\limits_{i \neq \xom}
 \mathbb{E}[T_{i,m}(n)] \theta_{\xom} + \sum\limits_{h \neq m} \mathbb{E}[T_{\xom,h}(n)] \theta_{\xom}
\end{equation}

From Theorem \ref{theorem:kth}, $T_{i,m}(n)$ and $T_{\aom,h}(n)$ are
bounded by
\begin{equation}
 \mathbb{E}[T_{i,m}(n)] \leq \frac{8 \ln n}{\Delta_{\xom, i}^2} + 1 + \frac{2
 \pi^2}{3},
\end{equation}
\begin{equation}
 \mathbb{E}[T_{\xom,h}(n)] \leq \frac{8 \ln n}{\Delta_{\xoh, \xom}^2} + 1 + \frac{2
 \pi^2}{3}.
\end{equation}
So,
\begin{equation}
\begin{split}
 \mathfrak{R}^\pi (\Theta,m;n) & \leq \sum\limits_{i \neq \xom}
 (\frac{8 \ln n}{\Delta_{\xom, i}^2} + 1 + \frac{2
 \pi^2}{3}) \theta_{\xom} \\
 & \quad + \sum\limits_{h \neq m} (\frac{8 \ln n}{\Delta_{\xoh, \xom}^2} + 1 + \frac{2
 \pi^2}{3}) \theta_{\xom}
\end{split}
\end{equation}

The upper bound for regret is:
\begin{equation}
\begin{split}
 & \mathfrak{R}^\pi (\Theta;n)  = \sum_{m = 1}^M \mathfrak{R}^\pi (\Theta,m;n) \\
 & \leq \sum_{m = 1}^M \sum\limits_{i \neq \xom}
 (\frac{8 \ln n}{\Delta_{\xom, i}^2} + 1 + \frac{2
 \pi^2}{3}) \theta_{\xom} \\
 & \quad + \sum_{m = 1}^M \sum\limits_{h \neq m} (\frac{8 \ln n}{\Delta_{\xoh, \xom}^2} + 1 + \frac{2
 \pi^2}{3}) \theta_{\xom}
\end{split}
\end{equation}

If we define $\Delta_{\min} = \min\limits_{1 \leq i \leq N, 1 \leq j
\leq M} \Delta_{i,j}$, and $\theta_{\max} = \max\limits_{1 \leq i
\leq N} \theta_{i}$, we could get a more concise (but looser) upper
bound as:
\begin{equation}
\begin{split}
 \mathfrak{R}^\pi (\Theta;n) \leq M(N+M-2)(\frac{8 \ln n}{\Delta_{\min}^2} + 1 + \frac{2
 \pi^2}{3}) \theta_{\max}.
\end{split}
\end{equation}
\end{IEEEproof}

Theorem \ref{theorem:distributed} shows that the regret of our DLP
algorithm is uniformly upper-bounded for all time $n$ by a function
that grows as $O(M(N+M)\ln n)$.

\section{Distributed Learning with Fairness} \label{sec:DLF}

\begin{figure}[h]
\centering
\includegraphics[width=0.24\textwidth]{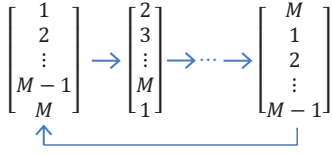}
\caption{Illustration of rotating the prioritization vector.}
\label{fig:1}
\end{figure}

For the purpose of fairness consideration, secondary users should be
treated equally, and there should be no prioritization for the
users. In this scenario, a naive algorithm is to apply Algorithm
\ref{alg:distributed} directly by rotating the prioritization as
shown in Figure \ref{fig:1}. Each user maintains two $1$ by $N$
vectors $(\hat{ \theta }_{j,i}^m)_{M \times N}$ and $(n_{j,i}^m)_{M
\times N}$, where the $j$-th row stores only the observation values
for the $j$-th prioritization vectors. This naive algorithm is shown
in Algorithm \ref{alg:d:naive}.

\begin{algorithm} [ht]
\caption{A Naive Algorithm for Distributed Learning Algorithm with
Fairness (DLF-Naive) Running at User $m$} \label{alg:d:naive}

\begin{algorithmic}[1]

\State At time $t$, run Algorithm \ref{alg:distributed} with
prioritization $K = ((m+t) \mod M) + 1$, then update the $K$-th row
of $(\hat{ \theta }_{j,i}^m)_{M \times N}$ and $(n_{j,i}^m)_{M
\times N}$ accordingly.

\end{algorithmic}
\end{algorithm}

We can see that the storage of Algorithm \ref{alg:d:naive} grows
linear in $MN$, instead of $N$. And it does not utilize the
observations under different allocation order, which will result a
worse regret as shown in the analysis of this section. To utilize
all the observations, we propose our distributed learning algorithm
with fairness (DLF) in Algorithm \ref{alg:d:fairness}.

\begin{algorithm} [ht]
\caption{Distributed Learning Algorithm with Fairness for $N$ Arms
with $M$ Users Running at User $m$ (DLF)} \label{alg:d:fairness}

\begin{algorithmic}[1]
\State \commnt{ Initialization}

\For {$t = 1$ to $N$} \label{a3:line:2}
    \State Play arm $k$ such that $k = ((m+t) \mod N) + 1$; \label{a3:line:3}
    \State $\hat{\theta}_{k}^m(t) = X_k(t)$;
    \State $n_{k}^m(t) = 1$;
\EndFor \label{a3:line:6}

\State \commnt{Main loop}

\While {1}
    \State $t = t + 1$;
    \State $K = ((m+t) \mod M) + 1$; \label{a3:line:9}
    \State Play an arm $k$ according to policy SL($K$) specified in Algorithm \ref{alg:kthLargest}; \label{a3:line:10}
    \State $\hat{\theta}_{k}^m(t) = \frac{\hat{\theta}_{k}^m(t-1) n_{k}^m(t-1) + X_{k}(t)}{ n_{k}^m(t-1) +1}$;
    \State $n_{k}^m(t) = n_{k}^m(t-1) + 1$;
\EndWhile
\end{algorithmic}
\end{algorithm}

Same as in Algorithm \ref{alg:distributed}, only two $1$ by $N$
vectors, $(\hat{\theta }_{i}^m)_{1 \times N}$ and $(n_{i}^m)_{1
\times N}$, are used by user $m$ to store the information after we
play an arm at each time slot.

Line \ref{a3:line:10} in Algorithm \ref{alg:d:fairness} means user
$m$ play the arm with the $K$-th largest expected reward with
Algorithm \ref{alg:kthLargest}, where the value of $K$ is calculated
in line \ref{a3:line:9} to ensure the desired arm to pick for each
user is different, and the users play arms from the estimated
largest to the estimated smallest in turns to ensure the fairness.


\theorem \label{theorem:d:naive} The expected regret under the
DLF-Naive policy specified in Algorithm \ref{alg:d:naive} is at most
\begin{equation}
\begin{split}
& \sum_{\xom \in \xO^*_m}\sum_{m = 1}^M \sum\limits_{i \neq \xom}
 (\frac{8 \ln \lceil n/M \rceil }{\Delta_{\xom, i}^2} + 1 + \frac{2
 \pi^2}{3}) \theta_{\aom} \\
 &+ \sum_{\xom \in \xO^*_m} \sum_{m = 1}^M \sum\limits_{h \neq m} (\frac{8 \ln \lceil n/M \rceil}{\Delta_{\xoh, \xom}^2} + 1 + \frac{2
 \pi^2}{3}) \theta_{\aom}.
\end{split}
\end{equation}

\begin{IEEEproof}
Theorem \ref{theorem:d:naive} is a direct conclusion from Theorem \ref{theorem:distributed} by replacing $n$ with $\lceil n/M \rceil$, and then take the sum over all $M$ best arms which are played in the algorithm.
\end{IEEEproof}

The above theorem shows that the regret of the DLF-Naive policy
grows as $O(M^2(N+M)\ln n)$.

\theorem \label{theorem:d:fairness} The expected regret under the
DLF policy specified in Algorithm \ref{alg:d:fairness} is at most
\begin{equation} \label{equ:theorem4}
\begin{split}
& M \sum\limits_{i = 1}^N
 (\frac{8 \ln n}{\Delta_{\min,i}^2} + 1 + \frac{2 \pi^2}{3}) \theta_{\max} \\
 & + M (M-1) \sum\limits_{i \in \xO_M^*} (\frac{8 \ln n}{\Delta_{\min,i}^2} + 1 + \frac{2 \pi^2}{3})
 \theta_{i},
\end{split}
\end{equation}
where $\Delta_{\min,i} = \min\limits_{1 \leq m \leq M}
\Delta_{\xom,i}$.

\begin{IEEEproof}


Denote $\kom(t)$ as the index of the arm with the $K$-th (got by
line \ref{a3:line:9} at time $t$ in Algorithm \ref{alg:d:fairness}
running at user $m$) largest expected reward. Denote $Q_i^m(n)$ as
the number of times that user $m$ pick arm $i \neq \kom(t)$ for $1
\leq t \leq n$.

We notice that for any arbitrary positive integer $l$ and any time
$t$, $Q_i^m(t) \geq l$ implies $n_i(t) \geq l$. So (\ref{equ:d002})
to (\ref{equ:p4}) in the proof of Theorem \ref{theorem:kth} still
hold by replacing $T_i(n)$ with $Q_i^m(n)$ and replacing $K$ with
$\kom(t)$. Note that since all the channels are with different
rewards, there is only one element in the set $\xA_{\kom(t)}$.

To find the upper bound of $\mathbb{E}[Q_i^m(n)]$, we should let $l$
to be $l \geq \left\lceil \frac{8 \ln n}{\Delta_{\min,i}^2}
\right\rceil$ such that (\ref{equ:de3}) and (\ref{equ:de6}) are
false for all $t$. So we have,
\begin{equation}
\begin{split} \label{equ:d07}
 & \mathbb{E}[Q_i^m(n)]
  \leq \left\lceil \frac{8 \ln n}{\Delta_{\min,i}^2}
   \right\rceil + \sum\limits_{t = 1}^{\infty} \sum\limits_{n_{j(t)} = 1}^{t-1} \sum\limits_{n_i = \left\lceil (8 \ln n)/
   \Delta_{\kom(t),i}^2
   \right\rceil}^{t-1} \\
   & (Pr\{\bX_{j(t), n_{j(t)}} \leq \theta_{j(t)} - C_{t, n_{j(t)}} \}  + Pr\{\bX_{i, n_i} \geq \theta_{i} + C_{t,n_i} \})\\
   & + \sum\limits_{t = 1}^{\infty} \sum\limits_{n_i = \left\lceil (8 \ln n)/
   \Delta_{\kom(t),i}^2
   \right\rceil}^{t-1} \sum\limits_{n_{h(t)} = 1}^{t-1} \\
   & ( Pr\{\bX_{i,
n_i} \leq \theta_{i} - C_{t, n_i}\} + Pr\{\bX_{h(t),
n_{h(t)}} \geq \theta_{h(t)} +C_{t, n_{h(t)}}\} )\\
& \leq \frac{8 \ln n}{\Delta_{\min,i}^2} + 1 + 2 \sum\limits_{t =
1}^{\infty}
 \sum\limits_{n_{j(t)} = 1}^{t-1}  \sum\limits_{n_i = 1}^{t-1} 2
 t^{-4} \\
 & \leq \frac{8 \ln n}{\Delta_{\min,i}^2} + 1 + \frac{2 \pi^2}{3}.
\end{split}
\end{equation}

Hence for user $m$, we could calculate the upper bound of regret considering the two possibilities as in the proof of Theorem \ref{theorem:distributed} as:
\begin{equation}
 \mathfrak{R}^\pi (\Theta,m;n) \leq \sum\limits_{i = 1}^N
 Q_i^m(n) \theta_{\max} + \sum\limits_{h \neq m} \sum\limits_{i \in \xO_M^*} Q_h^m(n) \theta_{i}
\end{equation}

So the upper bound for regret for $m$ users is:
\begin{equation}\label{equ:d10}
\begin{split}
 & \mathfrak{R}^\pi (\Theta;n)  = \sum_{m = 1}^M \mathfrak{R}^\pi (\Theta,m;n) \\
 & \leq  M \sum\limits_{i = 1}^N
 (\frac{8 \ln n}{\Delta_{\min,i}^2} + 1 + \frac{2 \pi^2}{3}) \theta_{\max} \\
 & + M (M-1) \sum\limits_{i \in \xO_M^*} (\frac{8 \ln n}{\Delta_{\min,i}^2} + 1 + \frac{2 \pi^2}{3}) \theta_{i}
\end{split}
\end{equation}

To be more concise, we could also write the above upper bound as:
\begin{equation}
\begin{split}
 \mathfrak{R}^\pi (\Theta;n) \leq M(N+M(M-1))(\frac{8 \ln n}{\Delta_{\min}} + 1 + \frac{2
 \pi^2}{3}) \theta_{\max}.
\end{split}
\end{equation}

\end{IEEEproof}

\theorem \label{theorem:d:f:nlarge} When time $n$ is large enough
such that
\begin{equation}
\frac{n}{\ln n} \geq \frac{8 (N+M)}{\Delta_{\min}^2} + (1 + \frac{2
\pi^2}{3})N + M,
\end{equation}
the expected regret under the DLF policy specified in Algorithm
\ref{alg:d:fairness} is at most
\begin{equation}
\begin{split}
& M \sum\limits_{i \notin \xO_M^*}
 (\frac{8 \ln n}{\Delta_{\min,i}^2} + 1 + \frac{2 \pi^2}{3}) \theta_{\max} + M^2 (1 + \frac{2 \pi^2}{3}) \theta_{\max}\\
 &  + M(M-1)(1 + \frac{2 \pi^2}{3})\sum\limits_{i \in \xO_M^*}\theta_{i}.
\end{split}
\end{equation}

\begin{IEEEproof}
The inequality (\ref{equ:d07}) implies that the total number of
times that the desired arms are picked by user $m$ at time $n$ is
lower bounded by $n - \sum\limits_{i = 1}^N (\frac{8 \ln
n}{\Delta_{\min,i}^2} + 1 + \frac{2 \pi^2}{3})$. Since all the arms
with $M$ largest expected rewards are picked in turn by the
algorithm, $\forall i \in \xO_M^*$, we have
\begin{equation}
n_i(n) \geq \frac{1}{M}\left(n - \sum\limits_{i = 1}^N (\frac{8 \ln
n}{\Delta_{\min,i}^2} + 1 + \frac{2 \pi^2}{3})\right).
\end{equation}
where $n_i(n)$ refers to the number of times that arm $i$ has been
observed up to time $n$ at user $m$. (For the purpose of simplicity,
we omit $m$ in the notation of $n_i$.)

Note that when $n$ is big enough such that $\frac{n}{\ln n} \geq
\frac{8 (N+M)}{\Delta_{\min}^2} + (1 + \frac{2 \pi^2}{3})N + M$, we
have,
\begin{equation}\label{equ:d08}
n_i(n) \geq \frac{1}{M}\left(n - \sum\limits_{i = 1}^N (\frac{8 \ln
n}{\Delta_{\min,i}^2} + 1 + \frac{2 \pi^2}{3})\right) \geq \lceil
\frac{8 \ln n}{\Delta_{\min}^2} \rceil.
\end{equation}

When (\ref{equ:d08}) holds, both (\ref{equ:de3}) and (\ref{equ:de6})
are false. Then $\forall i \in \xO_M^*$, when $n$ is large enough to
satisfy (\ref{equ:d08}),
\begin{equation}
\begin{split} \label{equ:d09}
 & \mathbb{E}[Q_i^m(n)] = \sum\limits_{t = N+1}^n \mathds{1} \{ I_i(t)\}\\
 & = \sum\limits_{t = N+1}^n \left( \mathds{1} \{ I_i(t) , \theta_i < \theta_K\} + \mathds{1} \{ I_i(t) , \theta_i > \theta_K
 \} \right)\\
 & \leq \sum\limits_{t = 1}^{\infty} \sum\limits_{n_{j(t)} = 1}^{t-1} \sum\limits_{n_i = \left\lceil (8 \ln n)/\Delta_{\min}^2
   \right\rceil}^{t-1} \\
   & (Pr\{\bX_{j(t), n_{j(t)}} \leq \theta_{j(t)} - C_{t, n_{j(t)}} \}  + Pr\{\bX_{i, n_i} \geq \theta_{i} + C_{t,n_i} \})\\
   & + \sum\limits_{t = 1}^{\infty} \sum\limits_{n_i = \left\lceil (8 \ln n)/
   \Delta_{\min}^2
   \right\rceil}^{t-1} \sum\limits_{n_{h(t)} = 1}^{t-1} \\
   & ( Pr\{\bX_{i,
n_i} \leq \theta_{i} - C_{t, n_i}\} + Pr\{\bX_{h(t),
n_{h(t)}} \geq \theta_{h(t)} +C_{t, n_{h(t)}}\} )\\
& \leq 1 + 2 \sum\limits_{t =
1}^{\infty}
 \sum\limits_{n_{j(t)} = 1}^{t-1}  \sum\limits_{n_i = 1}^{t-1} 2
 t^{-4} \leq 1 + \frac{2 \pi^2}{3}.
\end{split}
\end{equation}

So when (\ref{equ:d08}) is satisfied, a tighter bound for the regret
in (\ref{equ:theorem4}) is:
\begin{equation}\label{equ:d11}
\begin{split}
 & \mathfrak{R}^\pi (\Theta;n)  \leq M \sum\limits_{i \notin \xO_M^*}
 (\frac{8 \ln n}{\Delta_{\min,i}^2} + 1 + \frac{2 \pi^2}{3}) \theta_{\max} \\
 & \quad + M^2 (1 + \frac{2 \pi^2}{3}) \theta_{\max} + M(M-1)(1 + \frac{2 \pi^2}{3})\sum\limits_{i \in \xO_M^*}\theta_{i}.
\end{split}
\end{equation}
We could also write a concise (but looser) upper bound as:
\begin{equation}
\begin{split}
 \mathfrak{R}^\pi (\Theta;n) & \leq M(N-M)(\frac{8 \ln n}{\Delta_{\min}} + 1 + \frac{2
 \pi^2}{3}) \theta_{\max} \\
 & \quad + M^3 (1 + \frac{2 \pi^2}{3}) \theta_{\max}.
\end{split}
\end{equation}

\end{IEEEproof}

Comparing Theorem \ref{theorem:d:naive} with Theorem
\ref{theorem:d:fairness} and Theorem \ref{theorem:d:f:nlarge}, if we
define $C = \frac{8 (N+M)}{\Delta_{\min}^2} + (1 + \frac{2
\pi^2}{3})N + M$, we can see that the regret of the naive policy
DLF-Naive grows as $O(M^2(N+M)\ln n)$, while the regret of the DLF
policy grows as $O(M(N+M^2)\ln n)$ when $\frac{n}{\ln n} < C$,
$O(M(N-M)\ln n)$ when $\frac{n}{\ln n} \geq C$.

\section{Numerical Results}\label{sec:simulation}

\begin{figure*}[t]
    \centering
    \subfigure[$N=4$ channels, $M = 2$ secondary users, $\Theta = (0.9, 0.8, 0.7, 0.6)$.]
    {
        \includegraphics[width=0.31\textwidth]{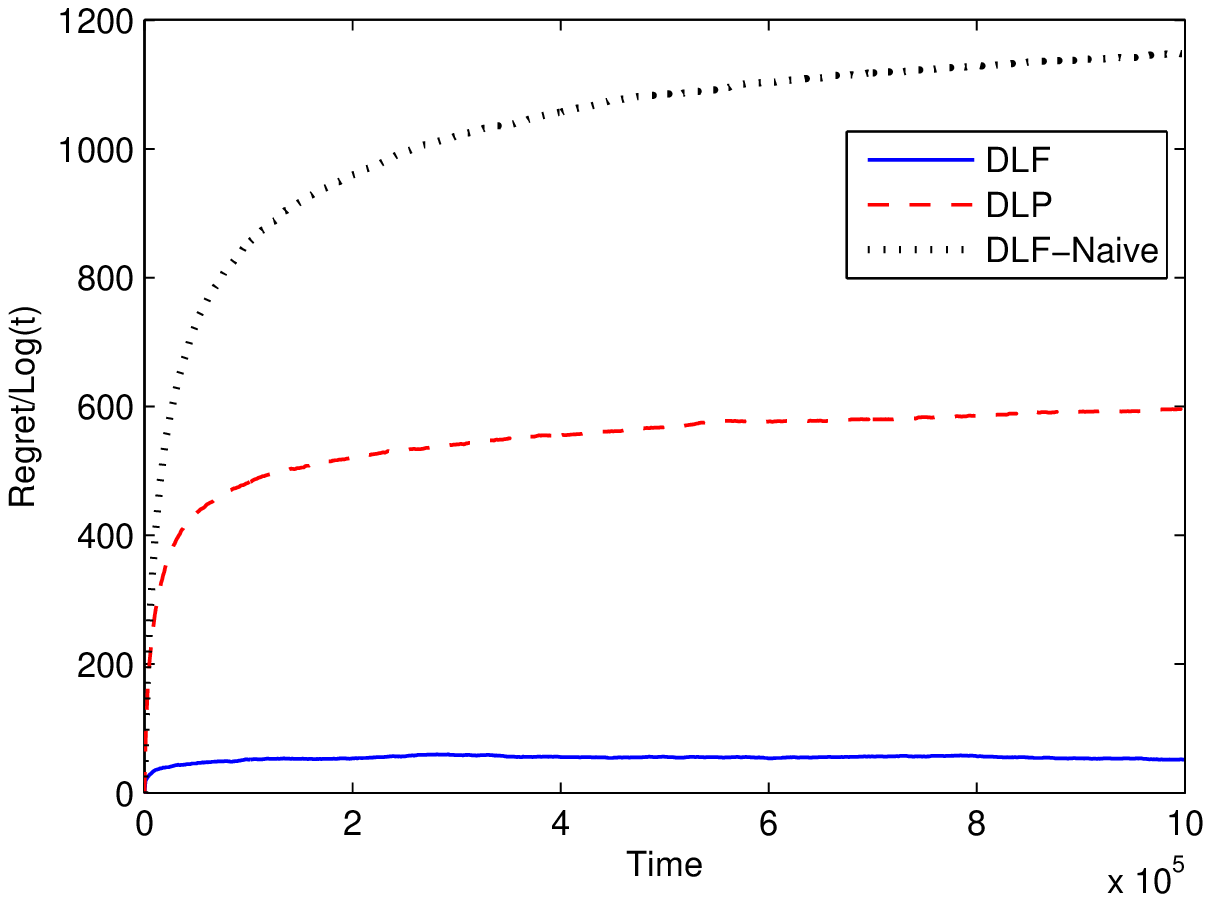}
        \label{fig:first_sub}
    }
    \subfigure[$N=5$ channels, $M = 3$ secondary users, $\Theta = (0.9, 0.8, 0.7, 0.6, 0.5)$.]
    {
        \includegraphics[width=0.31\textwidth]{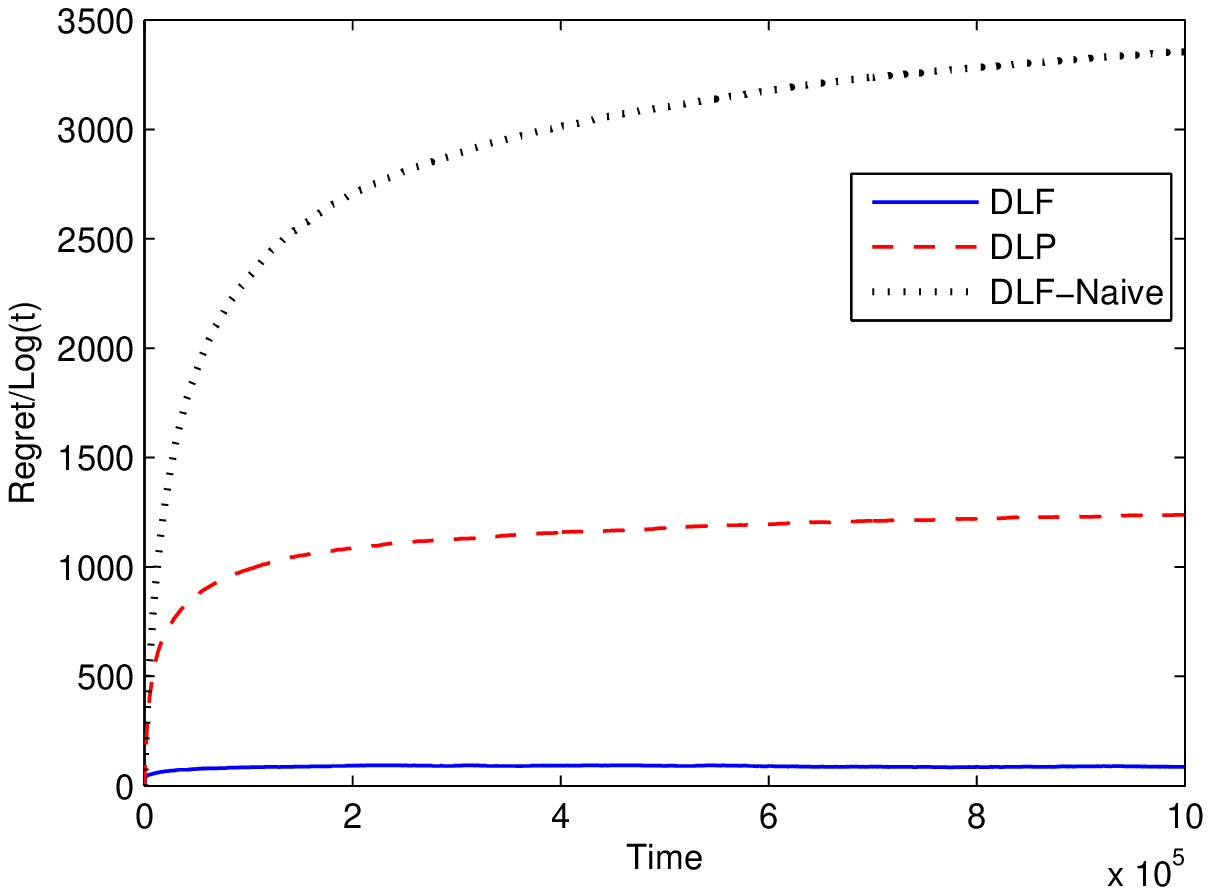}
        \label{fig:second_sub}
    }
    \subfigure[$N=7$ channels, $M = 4$ secondary users, $\Theta = (0.9, 0.8, 0.7, 0.6, 0.5, 0.4, 0.3)$.]
    {
        \includegraphics[width=0.31\textwidth]{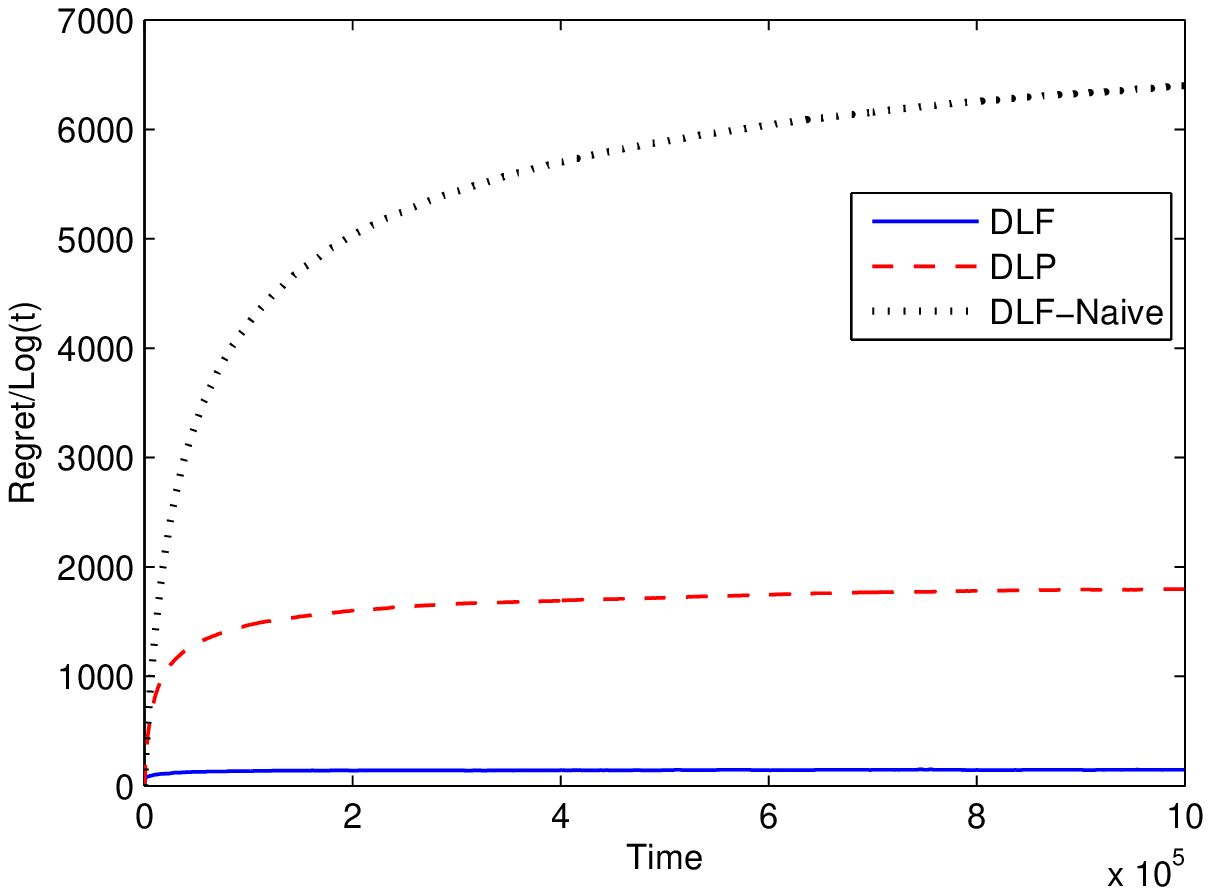}
        \label{fig:third_sub}
    }
    \caption{Normalized regret $\frac{\mathfrak{R}(n)}{\ln n}$ vs. $n$ time slots.}
    \label{fig:regretbound}
\end{figure*}

\begin{figure*}
    \centering
    \subfigure[DLP policy.]
    {
        \includegraphics[width=0.3\textwidth]{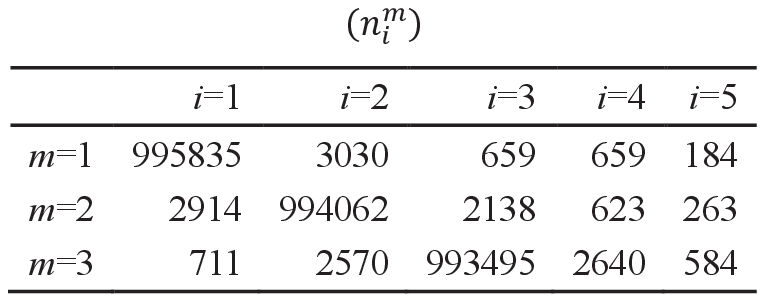}
        \label{fig:t2}
    }
    \subfigure[DLF policy.]
    {
        \includegraphics[width=0.3\textwidth]{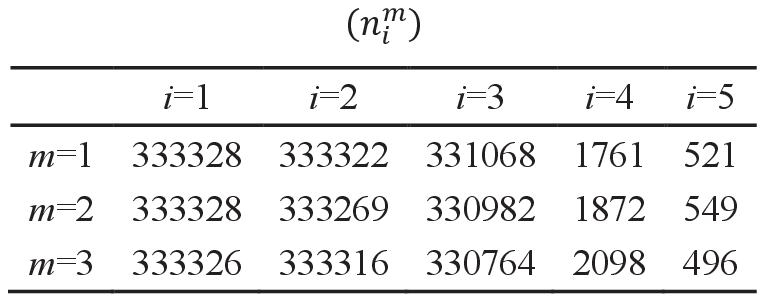}
        \label{fig:t1}
    }
    \\
    \centering
    \subfigure[DLF-Naive policy.]
    {
        \includegraphics[width=0.9\textwidth]{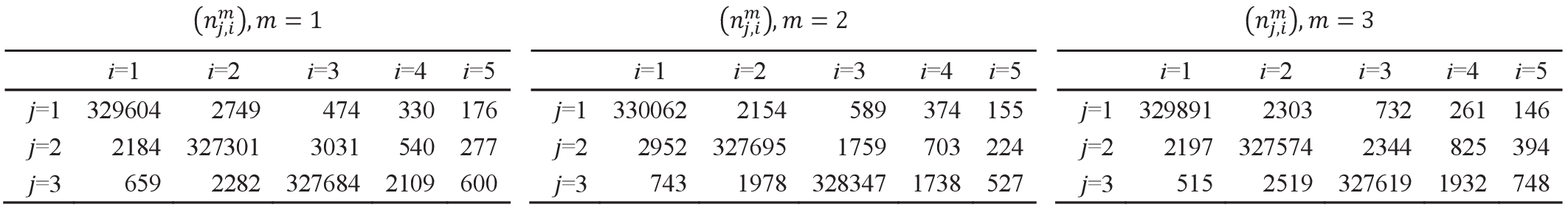}
        \label{fig:t3}
    }
    \caption{Number of times that channel $i$ has been chosen by user $m$ up to time $n = 10^6$, with $N=5$ channels, $M = 3$ secondary users and $\Theta = (0.9, 0.8, 0.7, 0.6, 0.5)$.}
    \label{fig:numberoftimes}
\end{figure*}

We present simulation results for the algorithms developed in this
work, varying the number of users and channels to verify the
performance of our proposed algorithms detailed earlier. In the
simulations, we assume channels are in either idle state (with
throughput 1) or busy state (with throughput 0). The state of each
$N$ channel evolves as an i.i.d. Bernoulli process across time
slots, with the parameter set $\Theta$ unknown to the $M$ users.

Figure \ref{fig:regretbound} shows the simulation results averaged
over 50 runs using the three algorithms, DLP, DLF-Naive, and DLF,
and the regrets are compared. Figure \ref{fig:first_sub} shows the
simulations for $N = 4$ channels, $M = 2$ users, with $\Theta =
(0.9, 0.8, 0.7, 0.6)$. In Figure \ref{fig:second_sub}, we have $N =
5$ channels, $M = 3$ users, and $\Theta = (0.9, 0.8, 0.7, 0.6,
0.5)$. In Figure \ref{fig:third_sub}, there are $N = 7$ channels,
and $M = 4$ users, with $\Theta = (0.9, 0.8, 0.7, 0.6, 0.5, 0.4,
0.3)$.

As expected, DLF has the least regret, since one of the key features
of DLF is that it does not favor any one user over another. The
chance for each user to use any one of the $M$ best channels are the
same. It utilizes its observations on all the $M$ best channels, and
thus makes less mistakes for exploring. DLF-Naive not only has the
greatest regret, also uses more storage. DLP has greater regret than
DLF since user $m$ has to spend time on exploring the $M-1$ channels
in the $M$ best channels expect channel $k \neq \xom$. Not only this
results in a loss of reward, this also results in the collisions
among users. To show this fact, we present the number of times that
a channel is accessed by all $M$ users up to time $n = 10^6$ in
Figure \ref{fig:numberoftimes}.

Figure \ref{fig:regretbound} also explores the impact of increasing
the number of channels $N$, and secondary users $M$ on the regret
experienced by the different policies with the minimum distance
between arms $\Delta_{\min}$ fixed. It is clearly that as the number
of channels and secondary users increases, the regret, as well as
the regret gap between different algorithms increases.




\section{Conclusion} \label{sec:conclusion}
The problem of distributed multi-armed bandits is a fundamental
extension of the classic online learning framework that finds
application in the context of opportunistic spectrum access for
cognitive radio networks. We have made two key algorithmic
contributions to this problem. For the case of prioritized users, we
presented the first distributed policy that yields logarithmic
regret over time without prior assumptions about the mean arm
rewards. For the case of fair access, we presented a policy that
yields order-optimal regret scaling in terms of the numbers of users
and arms, which is also an improvement over prior results.

Through simulations, we further show that the overall regret is lower for the fair access policy.
In future work, we plan to undertake more comprehensive simulation based comparison of
the proposed policy with previously proposed schemes, including over more realistic channel models.
We are also interested in considering extensions of our distributed policies to multi-armed
bandits with dependent arms, such as the combinatorial model considered in~\cite{Gai:2010}.


\begin{thebibliography}{1}

\bibitem{Yucek:2009}
T. Yucek and H. Arslan, ``A Survey of Spectrum Sensing Algorithms
for Cognitive Radio Applications'', \emph{IEEE Communications
Surveys \& Tutorials}, vol. 11, pp. 116-130, 2009.

\bibitem{Lai:Robbins}
T.~Lai and H.~Robbins, ``Asymptotically Efficient Adaptive
Allocation Rules'', \emph{Advances in Applied Mathematics}, vol. 6,
pp. 4-22, 1985.

\bibitem{Anantharam}
V. Anantharam, P. Varaiya, and J. Walrand, ``Asymptotically
Efficient Allocation Rules for the Multiarmed Bandit Problem with
Multiple Plays¡ª Part I: I.I.D. Rewards'', \emph{IEEE Transactions
on Automatic Control}, vol. 32, pp. 968-976, 1987.

\bibitem{Anantharam:1987}
V. Anantharam, P. Varaiya, and J. Walrand, ``Asymptotically
Efficient Allocation Rules for the Multiarmed Bandit Problem with
Multiple Plays - Part II: Markovian Rewards'', \emph{IEEE
Transactions on Automatic Control}, vol. 32, pp. 977-982, 1987.

\bibitem{Auer:2002}
P.~Auer, N.~Cesa-Bianchi, and P.~Fischer, ``Finite-time Analysis of
the Multiarmed Bandit Problem'', \emph{Machine Learning}, vol. 47,
pp. 235-256, 2002.

\bibitem{Gai:2010}
Y.~Gai, B.~Krishnamachari, and R.~Jain, ``Learning Multiuser Channel
Allocations in Cognitive Radio Networks: A Combinatorial Multi-armed
Bandit Formulation'', \emph{IEEE International Dynamic Spectrum
Access Networks (DySPAN) Symposium}, Singapore, April, 2010.

\bibitem{Liu:Zhao}
K.~Liu and Q.~Zhao, ``Distributed Learning in Multi-Armed Bandit
With Multiple Players'', \emph{IEEE Transactions on Signal
Processing}, vol. 58,  pp. 5667 - 5681, November, 2010.


\bibitem{Liu:zhao:2010}
K.~Liu and Q.~Zhao, ``Distributed Learning in Cognitive Radio
Networks: Multi-Armed Bandit with Distributed Multiple Players'',
\emph{Proc. of IEEE International Conference on Acoustics, Speech,
and Signal Processing (ICASSP)}, March, 2010.

\bibitem{Anandkumar:Infocom:2010}
A. Anandkumar, N. Michael, and A.K. Tang, ``Opportunistic Spectrum
Access with Multiple Users: Learning under Competition'',
\emph{Proc. of IEEE International Conference on Computer
Communications (INFOCOM)}, March, 2010.

\bibitem{Anandkumar:JSAC}
A. Anandkumar, N. Michael, A. Tang, and A. Swami, ``Distributed
learning and allocation of cognitive users with logarithmic
regret'', to appear in \emph{IEEE JSAC on Advances in Cognitive
Radio Networking and Communications}.

\bibitem{Pollard}
D.~Pollard, \textit{Convergence of Stochastic Processes}. Berlin:
Springer, 1984.

\end{thebibliography}
\end{document}